\documentclass{ifacconf}
\pdfoutput=1
\usepackage{graphicx}      
\usepackage{natbib}        

\usepackage{epsfig}  
\usepackage{amsmath, amssymb, epstopdf}
\usepackage[usenames, dvipsnames]{xcolor}
\usepackage{soul}
\usepackage{mwe, array, balance}
\usepackage{esvect, url, float, bm}
\usepackage{pgf,tikz, mathrsfs}
\usetikzlibrary{arrows}
\usepackage{multirow, bigstrut}
\usepackage{empheq}
\usepackage{enumerate, romannum, cancel}
\usepackage{tcolorbox,multicol, bigints}
\usepackage{subfigure}
\usepackage{enumitem}
\usepackage{parskip}
\makeatletter
\def\set@curr@file#1{%
	\begingroup
	\escapechar\m@ne
	\xdef\@curr@file{\expandafter\string\csname #1\endcsname}%
	\endgroup
}
\def\quote@name#1{"\quote@@name#1\@gobble""}
\def\quote@@name#1"{#1\quote@@name}
\def\unquote@name#1{\quote@@name#1\@gobble"}
\makeatother
\usepackage{graphics}
\graphicspath{{.}}

\newtheorem{remark}{Remark:}
\newtheorem{lemma}[thm]{Lemma}
\newtheorem{definition}{Definition: }

\newcommand{\mf}[1]{\mathbf{#1}}
\DeclareMathAlphabet{\pazocal}{OMS}{zplm}{m}{n}

\newcommand{\hatmap}[1]{{#1}^{\times}}
\newcommand{\veemap}[1]{{#1}^{\vee}}

\newcommand{\q}{q}
\newcommand{\dq}{\dot{q}}
\newcommand{\ddq}{\ddot{q}}

\newcommand{\inds}{\mathcal{S}}
\newcommand{\indi}{\mathcal{I}}
\newcommand{\indicator}[1]{\mathbf{1}_{\indi}(#1)}
\newcommand{\indicate}[1]{\mathbf{1}_{#1}}

\newcommand{\mbar}{\overline m}
\newcommand{\del}[1]{\delta#1}
\newcommand{\ez}{\bm{e_3}}
\newcommand{\currsys}{\emph{multiple quadrotors carrying a flexible hose}}
\newcommand{\Currsys}{\emph{Multiple quadrotors carrying a flexible hose}}

\newcommand{\referarxiv}[1]{See Appendix~#1}

\begin{document}
	\begin{frontmatter}
		
		\title{Multiple quadrotors carrying a flexible hose: dynamics, differential flatness, control\thanksref{footnoteinfo}} 
		
		\thanks[footnoteinfo]{This work was supported in part by {the UC Berkeley Fire Research Group (https://frg.berkeley.edu/), the College of Engineering} and Berkeley Deep Drive.}
		
		\author{Prasanth Kotaru,} 
		\author[Second]{Koushil Sreenath} 
		
		\address[Second]{Dept. of Mechanical Engineering, University of California, Berkeley, CA, 94720 (e-mail: \{prasanth.kotaru, koushils\}@berkeley.edu).}
		
		\begin{abstract}
	Using quadrotors UAVs for cooperative payload transportation using cables has been actively gaining interest in the recent years. Understanding the dynamics of these complex multi-agent systems would help towards designing safe and reliable systems. In this work, we study one such multi-agent system comprising of multiple quadrotors transporting a flexible hose. We model the hose as a series of smaller discrete links and derive a generalized coordinate-free dynamics for the same. We show that certain configurations of this under-actuated system are differentially-flat. We linearize the dynamics using variation-based linearization and present a linear time-varying LQR to track desired trajectories. Finally, we present numerical simulations to validate the dynamics, flatness and control. 
	
\end{abstract}

		\begin{keyword}
			Coordinate-free dynamics, variation based linearization, co-operative control, aerial manipulation, differential-flatness
		\end{keyword}
		
	\end{frontmatter}
	\setlength{\abovedisplayskip}{0pt}
	\setlength{\belowdisplayskip}{0pt}
	\setlength{\textfloatsep}{0.5em}
	
	\section{Introduction}
Aerial manipulation has been an active research area for many years now, due to the simplicity of the dynamics and control of multi-rotors. The ubiquity of these aerial vehicles resulted in their use in a wide range of applications. Few such applications include search and rescue [\cite{bernard2011autonomous}] and disaster management, {for instance,} using UAVs to monitor forest fires [\cite{merino2012unmanned}]. Payload delivery using aerial vehicles [\cite{googlewing}, \cite{amazonair}, \cite{palunko2012agile}] is another application that has earned much attention in the last few years. 

One extension of the payload carrying research is developing multi-rotor vehicles for active fire-fighting [\cite{aerones}] using a tethered hose that carries water and power. This enables carrying a fire hose to heights higher than a typical fire-truck ladder as well as fly longer due to the tethered power supply. Multi-rotors are also used to help string power cables between poles [\cite{skyscopes}], which typically is achieved using manned helicopters.  To achieve stable and safe control of these complex systems, it is important to understand the underlying governing principles and dynamics. In this work, we aim to model and control the dynamics of a multiple quadrotor system carrying a flexible cable/hose. 

\begin{figure}
	\centering
	\includegraphics[width=0.9\columnwidth]{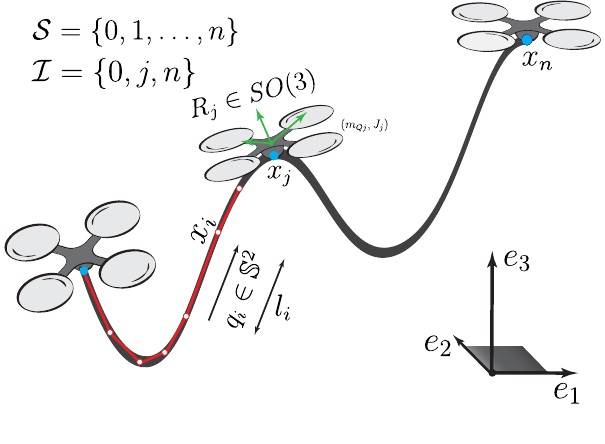}
	\caption{Multiple quadrotors carrying a flexible hose (with the hose modeled as a series of $n$ discrete-links). Links are massless with lumped mass at the ends and indexed through $\inds = \{0,1,...,n\}$. The set $\indi\subseteq\inds $ gives the set of indices where the hose is attached to a quadrotor. Each link is modeled as a unit-vector $q_i\in \mathbb{S}^2$. The configuration space of this system is $Q:= \mathbb{R}^3\times (\mathbb{S}^2)^n\times (SO(3))^{n_Q}$ $(n_Q=|\indi|)$.}
	\label{fig:intro}
\end{figure}

\subsection{Related Work}
There is a lot of literature on co-operative aerial manipulation, especially towards grasping and transporting payloads using multiple quadrotors [\cite{maza2009multi}, \cite{mellinger2013cooperative}, \cite{jiang2012inverse}, \cite{lee2017constraint}, \cite{michael2011cooperative}].  Trajectory tracking control for point-mass/rigid-body payloads suspended from multiple quadrotors is studied in [\cite{lee2013geometric}, \cite{goodarzi2016stabilization}, \cite{sreenath2013dynamics}, \cite{wu2014geometric}]. Similarly, for loads suspended using flexible cables, stabilizing controllers are presented in [\cite{goodarzi2015geometric}, \cite{goodarzi2015dynamics}] and these systems are shown to be differentially-flat in [\cite{kotaru2018differential}]. Tethered aerial vehicles have also been extensively studied in the literature, for instance stabilization of tethered quadrotor and nonlinear-observers for the same are discussed in [\cite{lupashin2013stabilization}, \cite{nicotra2014taut}, \cite{tognon2015nonlinear}]. Geometric control of a tethered quadrotor with a flexible tether is presented in [\cite{lee2015geometric}]. 

Most of the work discussed in the previous section models the tethers/cables either as rigid-links or as a series of links.
Partial differential equations have also been used to model a continuous mass system, such as the aerial refueling cable shown in [\cite{liu2017modeling}]. However, modeling the aerial cable as a finite-segment lumped mass [\cite{williams2007dynamics}, \cite{ro2010modeling}] is quite common in the literature due to the finite dimensionality of the state-space. However, most of these works assume Euler angles in the local frame to represent the attitude of the links. This results in complex equations of motion for the system that are also prone to singularities in case of aggressive motions. Therefore, in this work, we make use of coordinate-free representation that results in singularity-free and compact equations of motion.

\subsection{Challenges}

\Currsys~has multiple challenges in both modeling the dynamics and also designing a controller. Even though modeling the hose as a finite-segment lumped mass results in a finite-dimensional state space, it would still result in a large number of states depending on the choice of the number of discrete links. In addition, developing a controller is challenging due to the high under-actuation in the system. The swing of the cable, when not accounted for in the control, can have an adversarial effect. 

\subsection{Contributions}
In this paper, we build upon the work done in the literature to develop the dynamics and control of \currsys. This work is a step towards developing a system, with multiple quadrotor carrying a water-hose. However, for the purpose of this paper and as a first step, we consider no water flow in the hose. The contributions of this work are as follows, 
\begin{itemize}
	\item We derive a generalized coordinate-free dynamics for \currsys~ system. These dynamics can be extended to a tethered multiple quadrotor system.   
	\item We show that this system is \emph{differentially-flat} for certain configurations.  
	\item We present variation-based linearized dynamics and implement a time-varying LQR to track a time-varying desired trajectory. 
	\item Finally, we present numerical simulations to validate the dynamics and control. 
\end{itemize}
{To the best of authors knowledge, this is novel configuration of multiple quadrotors with a flexible hose and has not been studied prior to this work.}

\subsection{Organization}
Rest of the paper is organized as follows. Section~\ref{sec:dynamics} explains the system definition, notations and presents the derivation of the dynamics. In Section~\ref{sec:flatness} we show that the system~is differentially-flat. In section~\ref{sec:control}, we present a LQR control on the variation-linearized dynamics. Section~\ref{sec:simulations} presents numerical simulations validating the proposed controller. Finally, some of the limitations in this paper and potential directions to address them are discussed in Section~\ref{sec:discussion}. Concluding remarks are in Section~\ref{sec:conclusion}.

%
%

	\section{Dynamics}
\label{sec:dynamics}

Consider a flexible hose connected to multiple quadrotor UAVs as shown in Figure~\ref{fig:intro}. In this section, we present the coordinate-free dynamics for this system. We consider the following assumptions before proceeding to derive the dynamics:
\begin{enumerate}[label=A\arabic{*}.]
	\item No water/water-flow in the hose and thus also no pressure forces;
	\item Hose is modeled as a series of $n$ smaller links 
	connected by spherical joints;
	\item Each link is massless with lumped point-masses at the end with the hose mechanical properties like stiffness and torsional forces ignored.
	\item The quadrotors attach to the hose at their respective center-of-masses.
\end{enumerate}
In the following section, we present the notation used to describe the system. 

\subsection{Notation}
Dynamics for the model are defined using  geometric-representation of the states. Each link is a spherical-joint and is represented using a unit-vector $q\in\mathbb{S}^2:=\{x\in\mathbb{R}^3~|~\|x\|=1\}$. The position of one end of the cable is given in $\mathbb{R}^3$ and finally, the rotation matrix $R\in SO(3):=\{R\in \mathbb{R}^{3\times 3}|R^\top R = 1, det(R)=+1\}$ is used to represent the attitude of the quadrotor. 

Let the hose be discretized into $n$ links with the cable joints indexed as $\inds =\{0,1,\hdots,n\}$ as shown in Figure~\ref{fig:intro}. The position of one (starting) end of the hose is given as $x_0\in\mathbb{R}^3$ in the world-frame. The position of the link joints/point-masses is represented by $x_i\in\mathbb{R}^3$, where the link attitude between $x_{i{-}1}$ and $x_i$ is given by $q_i\in \mathbb{S}^2$ and length of this link-segment is $l_i$ \emph{i.e.,} $x_i = x_{i{-}1}{+
}l_iq_i$.  Also, $m_i$ is the mass of the lumped point-mass for link $i$. Let the set $\indi\subseteq \inds$ be the set of indices where the cable is attached to the quadrotor and $n_Q=|\indi|$ is the number of quadrotors. For the $j^{th}$ quadrotor, $R_j\in SO(3)$ is the attitude, $m_{Qj},J_j$ is its mass and inertia matrix (in body-frame) and $f_j\in\mathbb{R},M_j\in\mathbb{R}^3$ are the corresponding thrust and moment for all $j\in\indi$. Finally, the configuration space of this system is given as $Q:= \mathbb{R}^3\times (\mathbb{S}^2)^n\times (SO(3))^{n_Q}$. Table~\ref{tab:my_label} lists the various symbols used in this paper. 

\subsection{Derivation}
The kinematic relation between the different link positions is given using link attitudes as, 
\begin{align}
	x_i = x_0 + \sum_{k=1}^i l_k\q_k,& ~\forall~i\in \inds\backslash\{0\},\label{eq:link-positions}
\end{align}
and the corresponding velocities and accelerations are related as,
\begin{align}
	v_i = v_0 + \textstyle\sum_{k=1}^i l_k\dq_k,\quad 
	a_i = a_0 + \textstyle\sum_{k=1}^i l_k\ddot q_k.
\end{align}

Potential energy $\mathcal{U}:TQ\rightarrow \mathbb{R}$ of the system, due to hose and quadrotors' mass is computed as shown below,
\begin{align}
	\mathcal{U} = \sum_{i\in \inds}\mbar_ix_i\cdot g\bm{e_3},\label{eq:pe1}
\end{align}
where $\mbar_i = m_i  + m_{Qi}\indicate{i}$ is the net-mass at index $i$ and $\indicate{i}:=\indicator{i} = \begin{cases}
1,  ~ \textit{if } i\in \indi\\
0,  ~ \textit{else}
\end{cases} $ is an indicator function for the set $\indi$.

\begin{figure*}
	\normalsize
	\begin{tcolorbox}[title={Equations of motion for \currsys}]
		\begin{gather}
		\dot{x}_0 = v_0,~
		\dot{q}_i = \omega_i\times q_i, \label{eq:dyn_pos}\\
		\underbrace{\begin{bmatrix}
			M_{00}I_3 & -\hatmap{q}_1M_{01} & -\hatmap{q}_2M_{02} & \hdots &  -\hatmap{q}_nM_{0n} \\[2ex]
			-M_{10}\hatmap{q}_1 & -M_{11}I_3 & M_{12}\hatmap{q}_1\hatmap{q}_2 &{\scriptsize  \hdots} & M_{1n}\hatmap{q}_1\hatmap{q}_n\\[2ex]
			-M_{20}\hatmap{q}_2 & M_{21}\hatmap{q}_2\hatmap{q}_1 & -M_{22}I_3 & {\scriptsize \hdots} & M_{2n}\hatmap{q}_2\hatmap{q}_n\\[2ex]
			\vdots  & \vdots & \vdots & {\scriptsize \ddots} & \vdots  \\[2ex]
			-M_{n0}\hatmap{q}_n & M_{n1}\hatmap{q}_n\hatmap{q}_1 & M_{n2}\hatmap{q}_n\hatmap{q}_2 & {\scriptsize \hdots} & -M_{nn}I_3
			\end{bmatrix}}_{=:\mathbb{M}_{\{q_i\}}}
		\begin{bmatrix}
		\dot{v}_0 \\ \dot{\omega}_1 \\ \dot{\omega}_2 \\ \vdots \\ \dot{\omega}_n
		\end{bmatrix} =
		{\begin{bmatrix}
			\sum\limits_{i=1}^{n}M_{0i}\|\omega_i\|^2q_i + \sum_{k=0}^{n}u_k  \\[2ex] 
			-\sum\limits_{k=1}^{n}(M_{1k}\|\omega_k\|^2\hatmap{q}_1q_k){-}l_1\hatmap{q}_1\sum\limits_{k=1}^{n}u_k \\[2ex]	
			-\sum\limits_{k=1}^{n}(M_{2k}\|\omega_k\|^2\hatmap{q}_2q_k){-}l_2\hatmap{q}_2\sum\limits_{k=2}^{n}u_k \\[2ex]
			\vdots \\[2ex]
			-\sum\limits_{k=1}^{n}(M_{nk}\|\omega_k\|^2\hatmap{q}_nq_k){-}l_n\hatmap{q}_nu_n
			\end{bmatrix}}, \label{eq:dyn_q}\\
		\dot{R}_j = R_j\hatmap{\Omega}_j,~
		J_j\dot\Omega_j = M_j-\Omega_j\times J_i\Omega_j, \label{eq:dyn_R}
		\end{gather}
		$\forall i\in\inds\backslash\{0\}$, $j\in\indi$, $u_i = (- {\mbar_i}g\ez+ f_iR_i\ez\indicate{i} )$.
	\end{tcolorbox}
	\vspace*{4pt}
\end{figure*}
Kinetic energy $\mathcal{T}:TQ\rightarrow \mathbb{R}$ is similarly given as,
\begin{gather}
	\mathcal{T} = \sum_{i\in\inds}\frac{1}{2}\mbar_i\langle v_i,v_i\rangle + \sum_{j\in\indi} \frac{1}{2}\langle\Omega_j,J_j\Omega_j\rangle, \label{eq:ke1}
\end{gather}
where $\Omega_j$ is the angular velocity of the quadrotor $j$ in its body-frame. Dynamics of the system are derived using the Lagrangian method, where Lagrangian $\mathcal{L}:TQ\rightarrow \mathbb{R}$, is given as, 
\begin{gather*}
	\mathcal{L} = \mathcal{T}-\mathcal{U}. 
\end{gather*}

We derive the equations of motion using the Langrange-d'Alembert principle of least action, given below, 
\begin{align}
	\delta \int_{t_0}^{t_f}\mathcal{L}dt + \int_{t_0}^{t_f}\delta W_e dt = 0, \label{eq:leastaction}
\end{align}
where $\delta W_e$ is the infinitesimal work done by the external forces. $\delta W_e$ can be computed as,
\begin{align}
	\delta W_{e} &= \sum_{j\in\indi}\Big(\langle W_{1,j},\hat{M}_j \rangle + \langle W_{2,j}, f_jR_j\bm{e_3}\rangle \Big), \label{eq:dWe}
\end{align}
\begin{gather}
W_{1,j} =R^T_j\delta R_j, \\ 
W_{2,j} = \delta x_j = \delta x_0+\textstyle\sum_{k=1}^jl_k\delta q_k, 
\end{gather} are variational vector fields  [\cite{goodarzi2015geometric}] corresponding to quadrotor attitudes and positions. The infinitesimal variations on $q$ and $R$ are expressed as, 
\begin{gather*}
	\delta q = \hatmap{\xi}q = -\hatmap{q}\xi,~ \xi\in\mathbb{R}^3~\text{s.t.}~\xi\cdot q = 0, \\
	\del{\dq} = {-}\hatmap{q}\dot{\xi}{-}\hatmap{\dot{q}}\xi, \\
	\delta R  = R\hatmap{\eta},\quad \delta\hatmap{\Omega}=\hatmap{(\hatmap{\Omega}\eta)}{+}\hatmap{\dot \eta},\eta\in\mathbb{R}^3,
\end{gather*}
with the constraints $q\cdot\dot{q}=0$ and $q\cdot{\omega}=0$, $\omega$ is the angular velocity of $q$, s.t. $\dot{q} =\omega\times q$.  The cross-map is defined as $\hatmap{(\cdot)}:\mathbb{R}^3\rightarrow \mathbf{so(3)}$ \textit{s.t} $\hatmap{x}y=x\times y,\forall x,y\in\mathbb{R}^3$. Similarly, variations on the link positions are given as,
\begin{gather}
\del{x}_i = \del{x}_0 + \sum_{k=1}^i l_k\del{\q}_k =  \del{x}_0 {-} \sum_{k=1}^i l_k\hatmap{q}_k\xi_k,  \label{eq:delx}\end{gather}\begin{gather}
\del{v}_i = \del{v}_0{+}\sum_{k=1}^i l_k\del{\dq}_k = \del{v}_0{-}\sum_{k=1}^i l_k(\hatmap{q}_k\dot{\xi_k}{+}\hatmap{\dot{q}}_k\xi_k). \label{eq:delv}
\end{gather}
Finally, we obtain the equations of motion for the system by solving \eqref{eq:leastaction}.
\referarxiv{A}
for the detailed derivation. Equations of motion for the {\currsys} are given in \eqref{eq:dyn_pos}-\eqref{eq:dyn_R}. Note the mass-matrix $\mathbb{M}_{\{q_i\}}$ is a function of link attitudes $\{q_i\}=\{q_1,q_2,\hdots q_n \}$ and we use the following notation similar to [\cite{goodarzi2014geometric}]
\begin{gather}
M_{00}{=}\textstyle\sum_{k=0}^n \mbar_k,M_{0i}{=}l_i\textstyle\sum_{k=i}^n \mbar_k, \nonumber \\ M_{i0}{=}M_{0i},M_{ij} =\textstyle\sum_{k=\max\{ij\}}^n \mbar_k l_il_j.\label{eq:mass_var}
\end{gather}

\begin{remark}
	\label{remark:inputIndicator}
	In \eqref{eq:dyn_q}, note the use of $f_i, R_i$ for $i\notin \indi$, (since $i\notin \indi$ implies no quadrotor is attached at index $i$ and thus cannot have $f_i$ and $R_i$). However, this notation is used for convenience, since $i{\notin }\indi{\implies }\indicate{i}{=}0$ and thus $f_iR_i\ez\indicate{i}=0$, there by ensuring the right inputs to the system.
\end{remark}
\begin{remark}
	\label{remark:doua}
	Degrees of freedom for the \currsys~ is $DOF=3(n_Q{+}1){+}2n$ where $2n$ corresponds to the link attitudes DOF, $3n_Q$ the rotational DOF of the quadrotors and $3$ for the initial position $x_0$. Similarly, the degrees of actuation is $DOA=4n_Q$ corresponding to the $4$ inputs for each quadrotor. Thus, the degrees of under-actuation are $DOuA=2n+3-n_Q$. For a typical setup we have $n>>n_Q$, \emph{i.e.,} system is highly under-actuated.  
\end{remark} 
\begin{remark}
	\label{remark:tethered}
	For a tethered system, we can assume $x_0{\equiv}0$ $\forall~t$, i.e. the system is tethered to origin of the inertial frame, and we can derive the dynamics as earlier. Equations of motion for this system would be same as \eqref{eq:dyn_pos}-\eqref{eq:dyn_R}, without the equation corresponding to $\dot{v}_0$. 
\end{remark} 

\begin{table}[!t]
	\begin{center}
		\caption{List of various symbols used in this work. Note: $k{\in}\inds, i{\in}\inds\backslash\{0\}, j{\in}\indi$, WF - World frame, BF-Body-frame, $|\cdot |$ represents cardinality of a set.} \label{tab:my_label}
		\begin{tabular}{c|p{4cm}}
			Variables & Definition\\
			\hline\hline
			$n\in\mathbb{R}^+$ & Number of links in the hose.  \\
			$\mathcal{S}=\{0,1,\hdots,n\}$ & Set containing indices of the hose-segments.\\
			$x_k\in \mathbb{R}^3$ & Position of the $k^{th}$ point-mass of the hose in WF.\\
			$v_k\in \mathbb{R}^3$ & Velocity of the $k^{th}$ point-mass of the hose in WF.\\
			$l_i\in\mathbb{R}^+$ & Length of the $i^{th}$ segment.\\
			$m_k\in \mathbb{R}^+$ & Mass of the $k^{th}$ point-mass in the hose-segments.\\
			$\q_i\in\mathbb{S}^2$ & Orientation of the $i^{th}$ hose segment in WF. \\
			$\omega_i\in T_{q_i}\mathbb{S}^2$ & Angular velocity of the $i^{th}$ hose segment in WF. \\
			\hline\hline 
			$\mathcal{I}\subset \inds$ & Set of indices where the hose is
			attached to the quadrotor. \\
			$|\indi|=n_Q$& Number of quadrotors.\\
			$x_{Qj}\equiv x_j$ & Center-of-mass position of the $j^{th}$  quadrotor in WF. \\
			$R_j\in SO(3)$ & Attitude of the $j^{th}$  quadrotor w.r.t. WF.\\
			$\Omega_j\in T_{R_j}SO(3)$ & Angular velocity of the $j^{th}$ quadrotor in BF.\\
			$m_{Qj}$, $J_{j}$ & Mass \& inertia of the $j^{th}$ quadrotor.\\
			$f_j\in\mathbb{R},~M_j\in\mathbb{R}^3$ & Thrust and moment of the $j^{th}$ quadrotor in BF.\\
			\hline
			\hline 
			$\indicate{i} :=\indicator{i} = \begin{cases}
			1  ~ \textit{if } i\in \indi\\
			0  ~ \textit{else}
			\end{cases}$ & Indicator function for the set $\indi$.\\
			$\mbar_{k} = m_k + m_{Qk}\indicate{k}$  & Net mass at the $k^{th}$ link joint. \\
			$u_k = (- {\mbar_k}g\ez+ f_kR_k\ez\indicate{k} )$ & Net force due to thrusters \& gravity.\\
			\hline
		\end{tabular}
	\end{center}
\end{table}

	\section{Differential Flatness}
\label{sec:flatness}
\begin{figure*}
	\centering
	\includegraphics[width=0.8\textwidth]{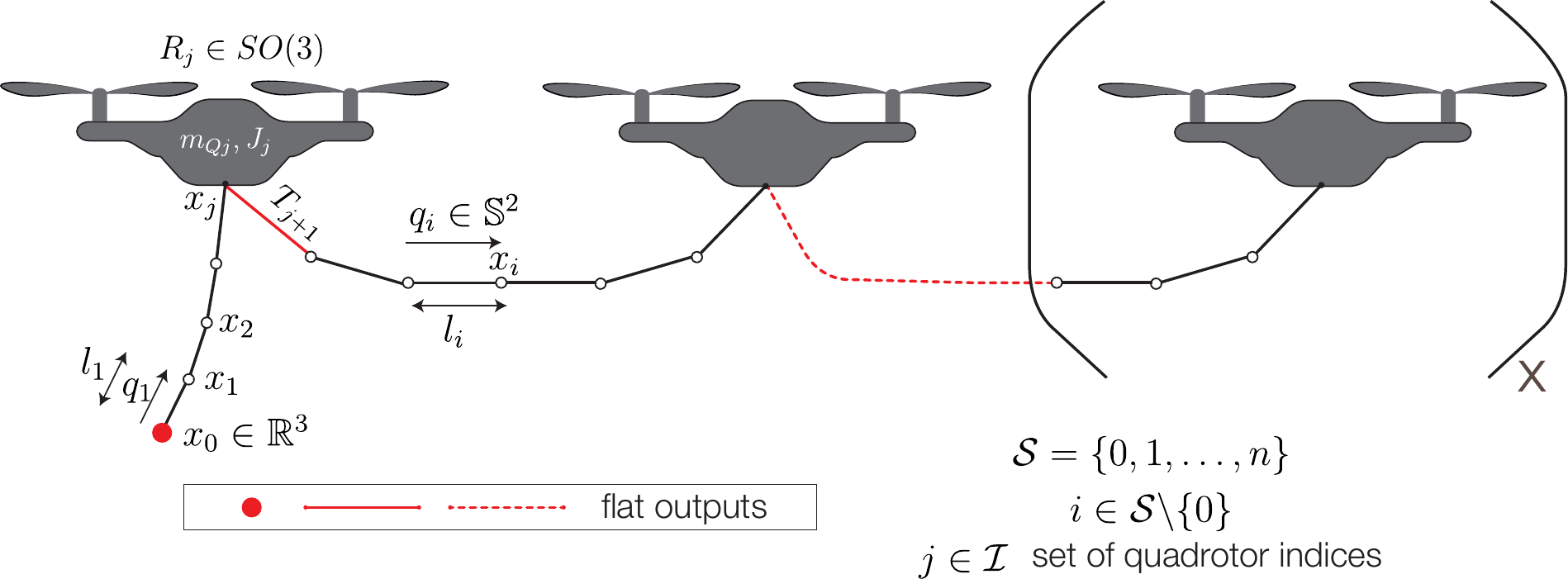}
	\caption{Configuration of the \currsys~illustrating the differential-flatness and its flat-outputs (shown in red).}
	\label{fig:flatness}
\end{figure*}
\begin{figure}
	\centering 
	\includegraphics[width=\columnwidth]{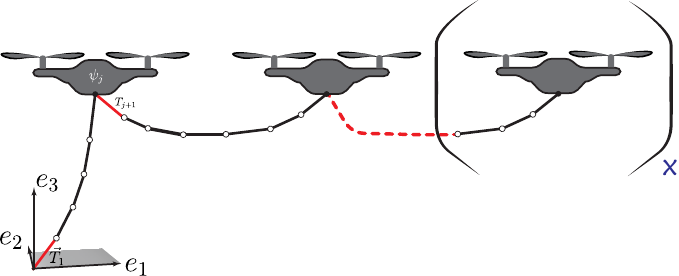}
	\caption{Tethered \currsys.}
	\label{fig:tethered}
\end{figure}
In the previous section, we derived the dynamics for \currsys. The system is under-actuated and thus the control of the system is challenging. In this section, we show that the system is differentially-flat. 
A nonlinear system is differentially-flat if a set of outputs of the system (equal to the number of inputs) and their derivatives can be used to determine the states and inputs without integration. 
\begin{definition}
	\label{defn:1}
	{\it Differentially-Flat System, [\cite{murray1995differential}]}: A system $\dot{\mathbf{x}} = f(\mf{x},\mf{u}),\,\mf{x} \in \mathbb{R}^n,\,\mf{u}\in \mathbb{R}^m,\,$ is differentially flat if there exists flat outputs $\mf{y}\in \mathbb{R}^m$ of the form $\mf{y} = \mf{y}(\mf{x},\mf{u},\dot{\mf{u}},\hdots,\mf{u}^{(\mf{p})})$ such that the states and the inputs can be expressed as $\mf{x} = \mf{x}(\mf{y},\dot{\mf{y}},\hdots,\mf{y}^{\mf{(q)}})$, $\mf{u} = \mf{u}(\mf{y},\dot{\mf{y}},\hdots,\mf{y}^{(\mf{q})})$, where $\mf{p},\,\mf{q}$ are positive finite integers.
\end{definition}
A quadrotor is a differentially-flat system with the quadrotor center-of-mass and yaw as the flat outputs [\cite{mellinger2011minimum}]. A quadrotor with cable suspended load (with the cable modeled as a massless link) is also shown to be differentially-flat with load position and quadrotor yaw as the flat outputs [\cite{sreenath2013geometric}]. 
Similarly, a quadrotor with flexible cable suspended load, with the cable modeled as series of smaller links is shown to be differentially-flat [\cite{kotaru2018differential}]. Again, here load position and quadrotor yaw are the flat-outputs. 
For the system defined in this work, the quadorotor-flexible cable segments are connected in series. Unlike previous work where each quadrotor has only one segment connected to it, each quadrotor in this system can have $0,1,$ or $2$ segments connected to it.  
In the following, we formalize the differential-flatness for certain configurations of the \currsys~system. 
\begin{lemma}
\label{lemma:diff}
${\pazocal{Y} = (x_0, \psi_j, T_{k{+}1})}~\forall j{\in}\indi~\&~ k{\in}\indi\backslash\{n\}$ are the set of flat-outputs for \currsys~with $n{\in}\indi$ (i.e., end of the cable is always attached to a quadrotor as shown in Figure~\ref{fig:flatness}), where $x_0{\in} \mathbb{R}^3$ is the position of the start of the cable, $\psi_j{\in}\mathbb{R}$ is the yaw angle of the $j^{th}$ quadrotor and $T_{k+1}\in \mathbb{R}^3$ is the tension vector in the $(k{+}1)^{th}$ link (as shown in Figure~\ref{fig:flatness}). 
\end{lemma}
\begin{pf}
\referarxiv{B} 
\end{pf}

\begin{remark}
	To determine the states and inputs of the system with $n-$ links, requires $(2n{+}4)$ derivatives of the flat-output $x_0$, $2^{nd}$ derivative of the yaw angle $\psi_j$ and $2(n{-}k){+}2$ derivatives of the tension vector $T_{k{+}1}$.
\end{remark}
\begin{cor}
	\label{cor}
	$\pazocal{Y} = (T_1, \psi_j, T_{k{+}1})~\forall j{\in}\indi~\&~ k{\in}\indi\backslash\{n\}$ are the flat-outputs for a tethered \currsys~shown in Figure~\ref{fig:tethered}, where $T_1 \in \mathbb{R}^3$ is the tension in the $1^{st}$ link, $\psi_j \in \mathbb{R}$ is the yaw angle of the quadrotor at index $j$ and $T_{k+1}\in\mathbb{R}^3$ is the tension vector in the $(k{+}1)^{th}$ link. 
\end{cor}
\begin{pf}
\referarxiv{B} 
\end{pf}

Differential-flatness is used in planning the system trajectories, where the flat outputs are used to plan in the lower-dimension space and the corresponding desired states and inputs are computed using differential flatness. In the next section, we present the linearized dynamics about any desired time-varying trajectory and use an LQR to track desired trajectories.

	\section{Control}
\label{sec:control}
Having presented differential-flatness in the previous section, we proceed to present control to track desired-trajectories generated using the flat-outputs in this section.
As presented in the Remark~\ref{remark:doua}, the given system is highly underactuated and thus controlling the system is challenging. In this section, we present a way to control the system by linearizing the dynamics in \eqref{eq:dyn_pos}-\eqref{eq:dyn_R} about a given desired time-varying trajectory\footnote{States \& inputs of the desired trajectories are represented with a subscript-$d$} $(x_{0d}(t), v_{0d}(t), q_{id}(t), \omega_{id}(t), R_{jd}(t), \Omega_{jd}(t))$, $\forall i\in \inds, j\in \indi$ and then implementing a linear controller. 

\subsection{Variation Based Linearization}
In this sub-section, we present the coordinate-free linear dynamics, obtained through variation based linearization of the nonlinear dynamics in \eqref{eq:dyn_pos}-\eqref{eq:dyn_R}. We use the variation linearization techniques described in [\cite{wu2015variation}] to obtain the linear dynamics. 
The error state of the linear-dynamics is given as,
	\begin{align}
		\mathbf{\delta x}  &= [ \delta x ,~\xi_1,~\hdots,~ \xi_n,~\delta v,~\delta\omega_1,~\hdots, \delta\omega_n, \nonumber \\
		& \quad \quad\eta_{j_1},~ \hdots,~ \eta_{j_{n_Q}},~ \delta\Omega_{j_1},~\hdots,~\delta\Omega_{j_{n_Q}}]^\top,
	\end{align}
	and the corresponding inputs as,
	\begin{align}
		\mathbf{\delta u} & = [\delta f_{j_1}, \delta f_{j_2}, \hdots, \delta f_{j_{n_Q}},  \delta M_{j_1}^\top , \delta M_{j_2}^\top,.., \delta M_{j_{n_Q}}^\top]^\top,
	\end{align}
	where $j_1, j_2,\hdots,j_{n_Q}$ are elements of $\indi$ arranged in increasing order.  The individual elements of the error state are computed as, 
	\begin{gather*}
		\del{x} = x{-}x_d,~\del{v}=v{-}v_d,\\
		\xi_i =\hatmap{q}_{id}q_i,~\del{\omega}_i=\omega_i-(-(\hatmap{q}_i)^2)\omega_{id},\\
		\eta_{j} = \frac{1}{2}\veemap{(R_{jd}^TR_j-R_j^TR_{jd})},~\del{\Omega_{j}} = \Omega_j - R_j^TR_{jd}\Omega_{jd}.
	\end{gather*} Finally, the linearized dynamics (\referarxiv{C} 
	for detailed derivation of the linearized dynamics) about a time-varying desired trajectory are given below, 
	\begin{gather}
	\delta\dot{\mathbf{ x}} = \mathcal{A}\delta \mathbf{ x} + \mathcal{B}\delta\mathbf{u}, \label{eq:var_lin_dyn}\\
	\mathcal{C}\delta\mathbf{x} = 0.\label{eq:var_lin_const}
	\end{gather}
	The linear dynamics matrices $\mathcal{A, B}$ are,{\scriptsize
	\begin{gather} 
		\mathcal{A}{=}\begin{bmatrix}
		0_{3,3} & 0_{3,3n}  & I_{3,3}  & 0_{3,3n}  & 0_{3,3n_Q}  & 0_{3,3n_Q}  \\
		0_{3n,3} & \alpha & 0_{3,3} & \beta & 0_{3,3n_Q}  & 0_{3,3n_Q} \\
		&&& \mathbb{M}_{\{q_{id}\}}^{-1}F &  \\
		0_{3n_Q,3} & 0_{3n_Q,3n} & 0_{3n_Q,3} & 0_{3n_Q,3n} & \gamma & I_{3n_Q,3n_Q}\\
		0_{3n_Q,3} & 0_{3n_Q,3n} & 0_{3n_Q,3} & 0_{3n_Q,3n} & 0_{3n_Q,3n_Q} & \nu
		\end{bmatrix}, \label{eq:var_linA}
	\end{gather} with,
	\begin{gather*}
		F = \begin{bmatrix} 
				O_{3,3} & [a]_i & O_{3,3} & [b]_i & [e]_j & O_{3n_Q, 3n_Q} \\
				O_{3n,3}& [c]_{i,j} & O_{3n,3} & [d]_{i,j} & [f]_{i,j}  & O_{3n_Q, 3n_Q}
			\end{bmatrix},
	\end{gather*}
	\begin{gather*}
		\alpha = \textit{bdiag}[q_{1d}q_{1d}^\top\hatmap{\omega}_{1d},~ q_{2d}q_{2d}^\top\hatmap{\omega}_{2d}, ~\hdots,~ q_{nd}q_{nd}^\top\hatmap{\omega}_{nd}],\\
		\beta = \textit{bdiag}[ \big(I_3{-}q_{1d}q_{1d}^\top\big), \big(I_3{-}q_{2d}q_{2d}^\top\big),\hdots, \big(I_3{-}q_{nd}q_{nd}^\top\big)],\\
		\gamma = \textit{bdiag}[-\hatmap{\Omega}_{j1d},~-\hatmap{\Omega}_{j2d},\hdots,-\hatmap{\Omega}_{jn_Qd}],\\
		\nu = \textit{bdiag}[J_1^{-1}(\hatmap{(J_1\Omega_{1d})}{-}\hatmap{\Omega}_{1d}J_1),\hdots,J_n^{-1}(\hatmap{(J_n\Omega_{nd})}{-}\hatmap{\Omega}_{nd}J_n)],
	\end{gather*}}
	and {\scriptsize
	\begin{gather}
		\mathcal{B} = \begin{bmatrix}
		O_{3(n{+}1), 4n_Q} \\ \mathbb{M}_{\{q_{id}\}}^{-1}G \\ 
		O_{3n_Q, 4n_Q} \\ 
		\begin{bmatrix}
		O_{3n_Q, n_Q} & \mu
		\end{bmatrix}
		\end{bmatrix},\text{ with }
		\mu = \textit{bdiag}[J^{-1}_{j1}, \hdots, J^{-1}_{jn_Q}],\label{eq:var_linB}
	\end{gather}
	\begin{gather*}
		G = \begin{bmatrix}
		\left.\begin{matrix}
		[g ]_i \\ [h]_{i,j}
		\end{matrix}\right| &  O_{(3(n+1), 3n_Q)}
		\end{bmatrix}.
	\end{gather*}}
	Next, the constraint matrix $\mathcal{C}$ is defined as,{\scriptsize 
	\begin{gather}
		\mathcal{C}{=}\begin{bmatrix}
			O_{n,3} & \mathcal{C}_1 & O_{n,3} & O_{n,3n} & O_{n,6n_Q} \\
			O_{n,3} & \mathcal{C}_2 & O_{n,3} & \mathcal{C}_1 &  O_{n,6n_Q}
		\end{bmatrix},
	\end{gather}	
	with
	\begin{gather*}
		\mathcal{C}_1 = bdiag(q^T_{1d}, q^T_{2d},\hdots,q^T_{nd}),
		\mathcal{C}_2 = bdiag({-}\omega^T_{1d}\hatmap{q}_{1d}, \hdots,{-}\omega^T_{nd}\hatmap{q}_{nd}).
	\end{gather*}}
	
	The rest of the elements are described below, {\scriptsize
	\begin{align*}
	a_i &= M_{0i}\big[(\hatmap{\dot{\omega}}_{id}-\|\omega_{id}\|^2I_3)\hatmap{q}_{id},\\
	b_i &= M_{0i}(2q_{id}\omega_{id}^\top),\quad i=\{1,\hdots,n\},
	\end{align*}
	\begin{align*}
	c_{ij} &= \begin{cases}
	\big[M_{io}\hatmap{\dot{v}_{0d}} -\sum_{j=1,j\neq i}^nM_{ij}\big(\hatmap{(\hatmap{q}_{jd})\dot{\omega}_{jd})} +\\ \quad \quad \quad \|\omega_{jd}\|^2\hatmap{q}_{jd} \big)- l_i{\big(\sum_{k=i}^n\hatmap{u}_k\big)} \big](-\hatmap{q}_i), &\quad i=j\\
	\big[M_{ij}\hatmap{q}_{id}\big(\hatmap{\dot\omega}_{jd} - \|\omega_{jd}\|^2I\big)\hatmap{q}_{jd}\big],& \quad i\neq j
	\end{cases}
	\end{align*}
	\begin{align*}
	d_{ij} &= \begin{cases}
	O_{3,3}, &\quad i=j\\
	M_{ij}\big[2\hatmap{q}_{id}q_{jd}\omega^\top_{jd}\big], &\quad i\neq j
	\end{cases}
	\end{align*}
	\begin{align*}
	e_j & =  -f_{jd}R_{jd}\hatmap{e}_3,\quad j\in{\indi}\\
	f_{ij} & = \begin{cases} \phi, & if~j\notin \indi \\
	-(l_i\hatmap{q}_i)f_{jd}R_{jd}\hatmap{e_{3}}, &if ~j\in\indi, j\geq i \\
	O_{3,3}, & if~ j\in \indi, j<i 
	\end{cases} \\
	g_j & = R_{jd}e_3,\\
	h_{ij} & = \begin{cases} \phi, & if~j\notin \indi \\
	(l_i\hatmap{q}_i)R_{jd}{e_{3}}, &if ~j\in\indi, j\geq i \\
	O_{3,1}, & if~ j\in \indi, j<i 
	\end{cases}
	\end{align*} }and \textit{bdiag} is block diagonal matrix. Note that $\mathbb{M}_{\{q_{id}\}}$ in \eqref{eq:var_linA}, \eqref{eq:var_linB} is the same mass matrix in \eqref{eq:dyn_q},  except is the function of desired link attitudes $\{q_{id}\}$. 
	
	As seen, \eqref{eq:var_lin_dyn}-\eqref{eq:var_lin_const} is a time-varying constrained linear system. The constraints arise due to the variation constraint on $S^2$ as discussed in [\cite{wu2015variation}]. Controllability of the constrained linear equation can be shown similar to [\cite{wu2015variation}], however, due the complexity of the matrices $\mathcal{A, B, C}$ computing the controllability matrix would be intractable. 
\subsection{Finite-Horizon LQR}

Assuming, we have the complete reference trajectory we can implement any linear control technique for \eqref{eq:var_lin_dyn}-\eqref{eq:var_lin_const}. Similar to [\cite[Lemma 1]{wu2015variation}], we can show that the constraint \eqref{eq:var_lin_const} is time-invariant, i.e., if the initial condition satisfies the constraint, solution to the linear system would satisfy the constraint for all time. However, due to this constraint, the controllability  matrix computed using $\mathcal{A, B}$ might not be full-rank and requires state transformation into the unconstrained space to result in full-rank controllability matrix. 

Instead, we opt for a finite-horizon LQR controller for the variation-linearized dynamics about a time-varying desired trajectory. We chose a finite-time horizon $T$, the terminal cost matrix $P_T$ and pick cost matrices for states $Q_1=Q_1^T$ and inputs $Q_2=Q_2^T$. Finally, we solve the continuous-time Ricatti equation backwards in time to obtain the gain matrix $P(t)$, that satisfies,
\begin{gather}
	-\dot{P} = Q_1-P\mathcal{B}Q_2^{-1}\mathcal{B}^TP + \mathcal{A}^TP + P\mathcal{A}.
\end{gather} 
The above equation is solved offline and stored in a table for online computation. Note that the explicit time dependence of $P, \mathcal{A}, \mathcal{B}$ is dropped for convenience. Finally, the feedback gain for the control input is computed as, 
\begin{gather}
K = R^{-1}\mathcal{B}P,\quad \del{\mathbf{ u}} = -K\del{\mathbf{ x}}.
\end{gather}
Since the gains are computed backwards in time, the computed input would result in a stable control for the constrained linear-system. The net control-input to the nonlinear system can be compute as, 
\begin{gather}
	u(t) = u_d(t) + \del{\mathbf{ u}}.
\end{gather} In the next section, we present few numerical simulations with the finite-horizon LQR performing tracking control on the full nonlinear-dynamics.

	\section{Numerical Simulations}
\label{sec:simulations}
In this section, we present numerical results to validate the dynamics and control discussed in the earlier sections. We present numerical simulations for tracking control for a desired setpoint and circular trajectory.\footnote{MATLAB code for the simulations can be found at \url{https://github.com/HybridRobotics/multiple-quadrotor-flexible-hose}. Video for simulations is at \url{https://youtu.be/i3egJ4fcAKM}. } 
\subsection{Setpoint Tracking}
\label{sec:setpoint}
\subsubsection{(i). Two Quadrotor system:} 
\begin{figure}
	\centering
	\includegraphics[width=\columnwidth]{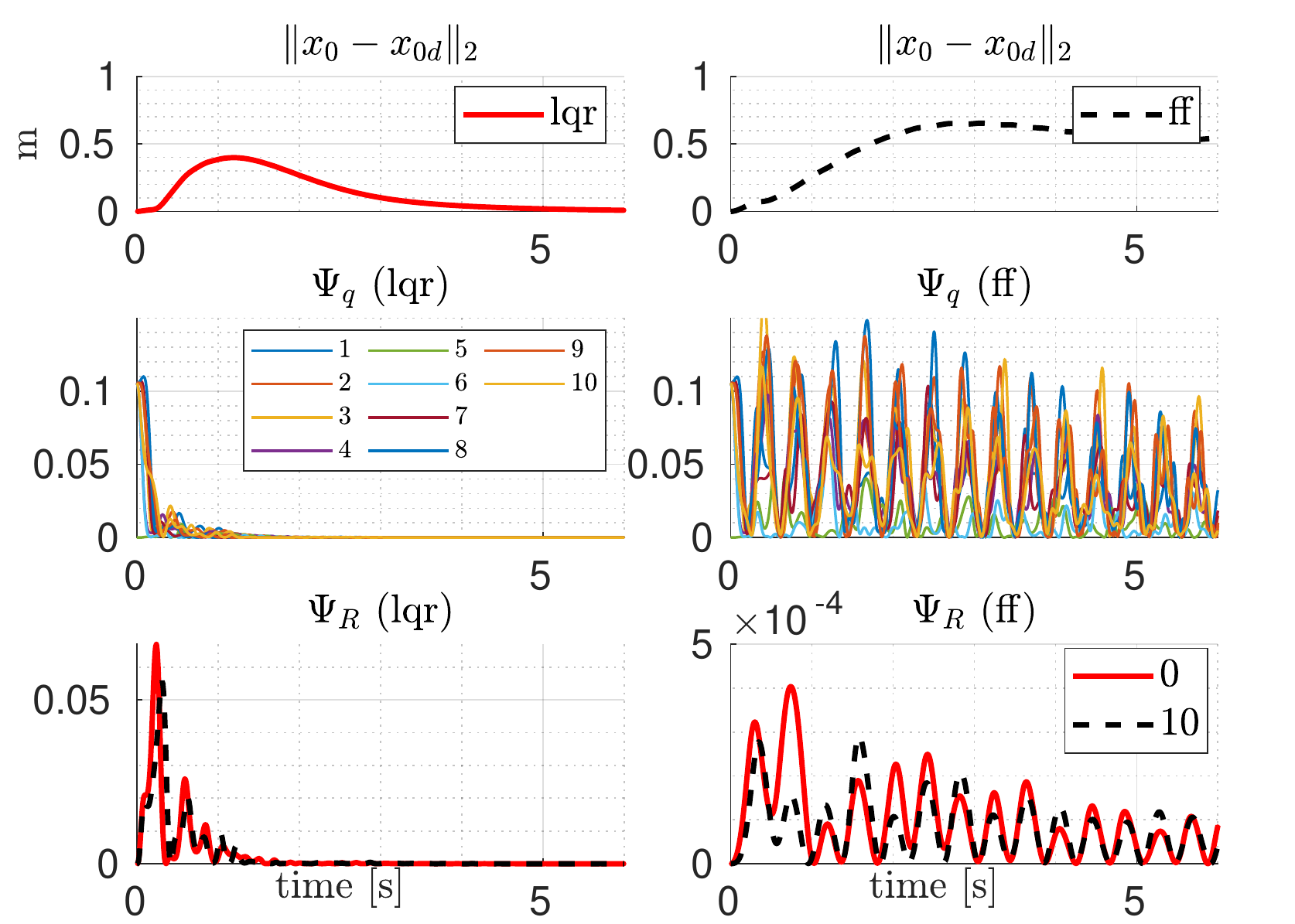}
	\caption{List of errors comparing the LQR control on the whole system (lqr) and feed-forward control on the quadrotor-position (ff). $\Psi_q$ is the hose link attitude errors as defined in \eqref{eq:PSIq} and $\Psi_R$ is the quadrotor attitude configuration error defined in \eqref{eq:PSIR}.}
	\label{fig:sec1_setpoint}
\end{figure}

Following parameters are considered for the simulations, 
\begin{gather*}
n= 10,~n_Q=2,~ \indi=\{ 0, 10\},~m_i=0.0909kg,~l_i=0.1m,\\
m_{Qj}=0.85kg,~J_j=diag([.0557, .0557, 0.1050])kgm^2
\end{gather*}
and the setpoint is given as, 
\begin{gather*}
x_{0d} = [0,0,0]^T, x_{nd}=[0.6,0.0,0.0]^T,
\end{gather*}
with the cable hanging between these two points. Degrees of freedom and under-actuation for this setup are $\#DOF=29,~\#DOuA=21$ respectively. The linear dynamics $\mathcal{A, B, C}$ are computed about this setpoint $x_d$. Here, we compare two different controllers, $(i)$ the finite-horizon LQR discussed in the previous-section and $(ii)$ position-controllers on the two quadrotors with feed-forward forces due to the cable at steady state. We start with some initial error in the cable orientation and the resulting error plots are shown in Figure~\ref{fig:sec1_setpoint}. As seen in the Figure, errors for cable position $x_0$, cable attitudes and quadrotors' attitude converge to origin. Attitude errors for the hose links is defined as the configuration error on $S^2$, 
\begin{gather}
\Psi_q=1-q^T_{id}q_i, \label{eq:PSIq}
\end{gather}
and similar quadrotor attitude error is defined as,
\begin{gather}
	\Psi_R=0.5Tr(I-R^T_{jd}R_{j}). \label{eq:PSIR}
\end{gather}
For the position control with feed-forward forces, even though the quadrotor attitudes are zero, the initial error in cable orientation results in oscillations in the cable. These oscillations are not accounted for in the control and can be seen in Figure~\ref{fig:sec1_setpoint}. 
%
%
\subsubsection{(ii). Three Quadrotor system:} 
\begin{figure}
	\centering
	\includegraphics[width=\columnwidth]{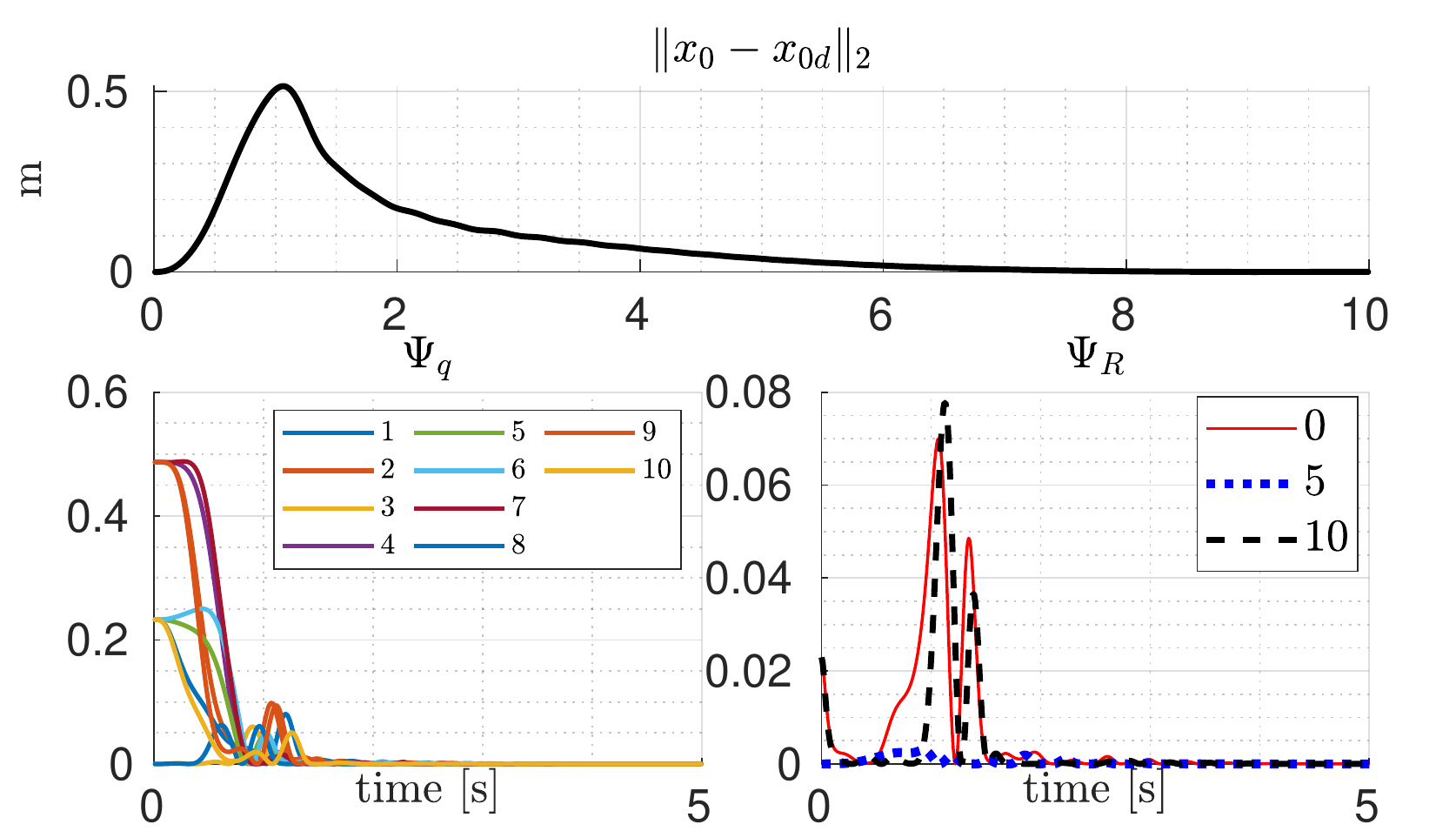}
	\caption{Tracking errors for desired set-point for $n_Q=3$ with the LQR control. $\Psi_q,\Psi_R$ are as defined in \eqref{eq:PSIq}-\eqref{eq:PSIR}.}
	\label{fig:sec1c_errors}
\end{figure}

\begin{figure*}
		\centering
		\includegraphics[width=1\textwidth]{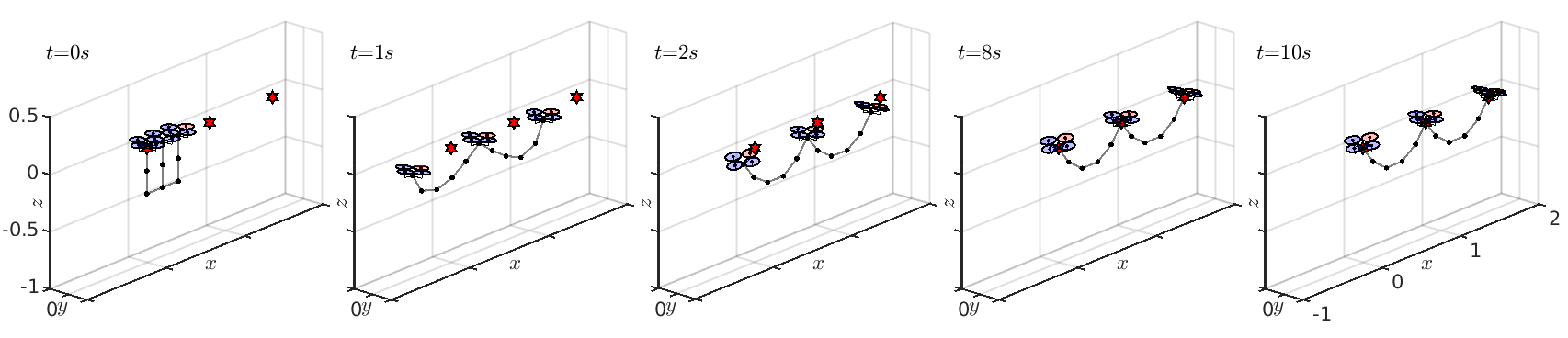}
	\caption{Snapshots of 3 quadrotor-10 link system while tracking a setpoint. Setpoints for the quadrotor position is shown by the red-hexagrams.}
	\label{fig:sec1c_traj}
\end{figure*}

Setpoint tracking for cable suspended from three-quadrotors is presented here. Parameters for the system are as follows,
\begin{gather*}
n= 10,~n_Q=3,~ \indi=\{ 0, 5, 10\},~m_i=0.0909kg,~l_i=0.2m,
\end{gather*}
and $\#DOF=32, \#DOuA=20$. Various tracking errors for the system are presented in Figure~\ref{fig:sec1c_errors} and snapshots for the system are shown in Figure~\ref{fig:sec1c_traj}. 

\subsection{Trajectory Tracking}
\begin{figure}
	\centering
	\includegraphics[width=\columnwidth]{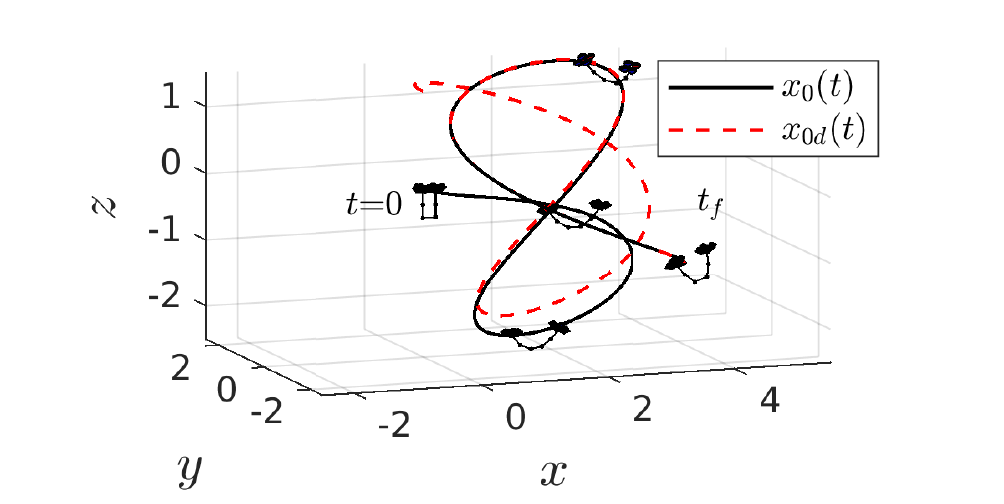}
	\caption{Snapshots of the \currsys~system while tracking the desired trajectory (shown in red) and the resulting trajectory when started with an initial error(shown in black).}
	\label{fig:sec2_traj}
\end{figure}

In this section, we show that the presented controller tracks a desired time-varying trajectory with initial errors. We use the following system parameters, 
\begin{gather*}
n= 5,~n_Q=2, \indi=\{ 0, 5\},~m_i=0.1667kg,~l_i=0figures.2m
\end{gather*}
and the rest same as those given in Section~\ref{sec:setpoint}. We consider the following flat output trajectory,
\begin{gather*}
x_0= \begin{bmatrix}
a_x(1{-}\cos(2f_1\pi t))\\
a_y\sin(2f_2\pi t)\\
a_z\cos(2f_3\pi t)
\end{bmatrix},~
\bar{T}_1 = \begin{bmatrix}
2.74 \\ 0.0 \\ -3.27
\end{bmatrix},~
\psi_{0}\equiv \psi_{5} \equiv 0,\\
f_1 = \frac{1}{4}, f_2 = \frac{1}{5}, f_3 = \frac{1}{7},
a_x = 2, a_y = 2.5, a_z = 1.5.
\end{gather*}
Rest of the states and inputs can be computed using differential-flatness.
We use the linearized-dynamics and the finite-horizon LQR presented in the previous sections to achieve the tracking control. Following weights are used for the LQR, 
\begin{gather*}
Q_x = 0.5I_6, ~Q_q = 0.75I_{6n},~Q_R = I_{3n_Q}, Q_\Omega = 0.75I_{3n_Q},\\
Q = bdiag(Q_x, Q_q, Q_R, Q_{\Omega}),\\
R  = 0.2I_{4n_Q},
P_T = 0.01I_{n_x},
\end{gather*}
where $n_x = 6{+}6n{+}6n_Q$. Figure~\ref{fig:sec2_traj} shows snapshots of the system at different instants along the trajectory. The proposed controller tracks the desired trajectory (shown in red) when started with an initial error. 

\begin{figure}
	\centering
	\includegraphics[width=\columnwidth]{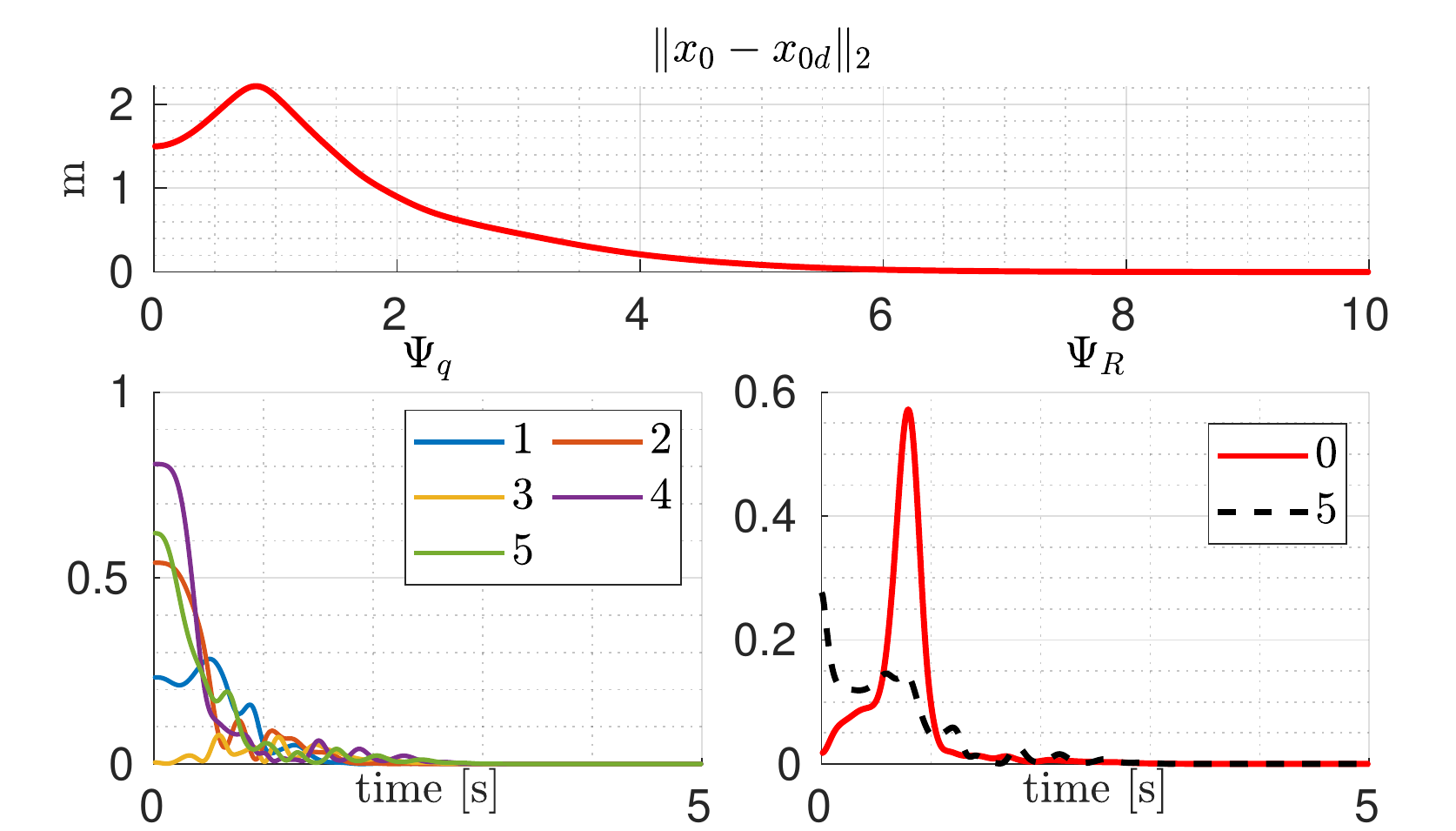}
	\caption{Errors for the trajectory tracking control shown in Fig.~\ref{fig:sec2_traj}. $\Psi_q,\Psi_R$ are as defined in \eqref{eq:PSIq}-\eqref{eq:PSIR}.}
	\label{fig:sec2_errors}
\end{figure}


	\section{Results and Discussion}
\label{sec:discussion}
Having presented numerical results to validate our controller, we now present some discussion on limitations and future work. 

\subsection*{Limitations}
\begin{figure}
	\centering
	\includegraphics[width=\columnwidth]{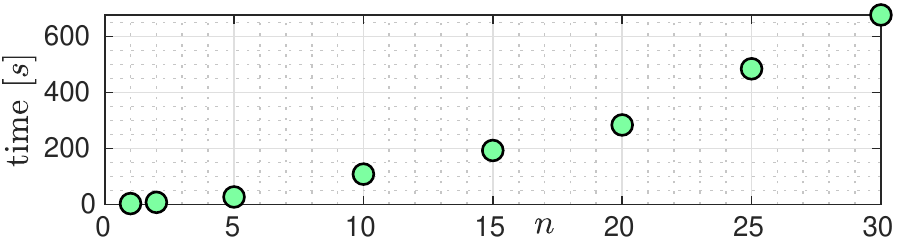}
	\caption{Computation time to simulate $10s$ for different discretizations $n$. }
	\label{fig:simtime}
\end{figure}

Though increasing discretization  helps better represent the dynamics of an hose system, it also increases the computation-complexity. To better study the effect of discretization we ran multiple simulations with different discretizations for a fixed cable length and mass. We used only control on the quadrotor-positions with feed-forward cable tensions. We used MATLAB 2018a with Intel Core $i7{-}6850\textsc{K cpu}@3.60GHz\times 12$ to run the simulations. Computation times to simulate $10s$ for different $n$ are shown in Figure~\ref{fig:simtime}. As illustrated in the Figure, computation time increases super-linearly with $n$. 

{While differential-flatness can be used to plan trajectories in the flat-output space and compute desired states and inputs, this computation requires computing $q_i$ and its derivatives from tension $\vec{T}_i$ and its derivatives, i.e. $q_i=\frac{\vec{T}_i}{\|\vec{T}_i\|}$. The complexity of this computation increases for higher-derivatives.} 

In addition, as listed in the Section~\ref{sec:dynamics}, we don't consider the mechanical properties of the hose when deriving the dynamics. Thus, the dynamics derived and the subsequent presented control might not completely capture the system fully and might lead to instability in cases when hose properties are important, such as when water flows in the hose. 

\subsection*{Future Work}
As part of future work, we would like to address some of the limitations listed in the previous sections, such as, \emph{(i)} number of discretizations, \emph{(ii)} number of derivatives to computed, and \emph{(iii)} water flow in the hose. We would like to study the current system along with all the mechanical properties of the cable and develop controllers for such systems. In addition, to implement the control we require state estimation of the cable which as shown is modeled as $(S^2)^n$. Towards this end we hypothesize [\cite{kotaru2019variation}] can be extended to estimate the cable state. A possible method to improve the computation time would be to use limited cable states like mid-position of the cable etc., to develop a controller.

\section{Conclusion}
\label{sec:conclusion}
In this work we have studied the \currsys~system. We modeled the flexible-hose as a series of smaller discrete-links with lumped mass and derived the coordinate-free dynamics using Langrange-d'Alembert's principle. We also showed that the given system is differentially-flat, as long as the end of the hose is connected to a quadrotor. Variation-based linearized dynamics were derived about time-varying desired trajectory. We showed tracking control for the system using finite-horizon LQR for the linear dynamics and validated this through numerical simulations with up-to 10 discretizations of the hose. Finally, we discussed some of the limitations due to the assumptions and directions for future work.

	\balance
	\bibliographystyle{ifacconf}
	\bibliography{references}
	
\newpage
\appendix
\section{Dynamics Derivation}
\label{sec:AppendixA}
In this section, we present the detailed derivation of the equations of motion, \eqref{eq:dyn_pos}-\eqref{eq:dyn_R}, for the given system. 
Starting with the  principle of least action in \eqref{eq:leastaction} and substituting for Lagrangian and virtual work from \eqref{eq:pe1},\eqref{eq:ke1} and \eqref{eq:dWe}, we have,
\begin{gather}
\delta\int\Big( \sum_{i\in\inds}\frac{1}{2}\mbar_i\langle v_i,v_i\rangle -\mbar_ig\ez\cdot x_i+ \sum_{j\in\indi} \underbrace{\frac{1}{2}\langle\Omega_j,J_j\Omega_j\rangle}_{rot.energy}\Big) \nonumber \\ 
\int\Big(\sum_{j\in\indi}\underbrace{\langle W_{1j},\hat{M}_j \rangle}_{rot.work} + \langle W_{2j}, f_jR_j\bm{e_3}\rangle \Big)dt.
\end{gather}
Separating and solving the rotational components, we have the following rotational dynamics
\begin{align}
J_j\dot{\Omega}_j = M-\Omega_j\times J_j\Omega_j,\quad \forall ~j\in \indi
\end{align}
Taking variation on rest of the equation results in,
\begin{align*}
\int \sum_{i\in \inds} \Bigg[{\mbar_i}\langle \del{v}_i,v_i\rangle + \del{x}_i\cdot \underbrace{(- {\mbar_i}g\ez+ f_iR_i\ez\indicate{i} )}_{u_i}\Bigg]dt = 0.
\end{align*}
Expanding the summation,
\begin{align*}
\bigint \begin{pmatrix}
{\mbar_0}\langle \del{v}_0,v_0\rangle + \del{x}_0\cdot u_0 + \\
{\mbar_1}\langle \del{v}_1,v_1\rangle + \del{x}_1\cdot u_1 + \\
\vdots \\
{\mbar_n}\langle \del{v}_n,v_n\rangle + \del{x}_n\cdot u_n 
\end{pmatrix}dt = 0
\end{align*}
Replacing the variations $\del{v}_i,\del{x}_i$ with their expansions \eqref{eq:delx}, \eqref{eq:delv}, we get,
{\small \begin{align}
\bigint \begin{pmatrix}
\big({\mbar_0}\langle \del{v}_0,v_0\rangle + \del{x}_0\cdot u_0\big) + \\
\big({\mbar_1}\langle \del{v}_0 {-} l_1(\hatmap{q}_1\dot{\xi_1}+\hatmap{\dot{q_1}}\xi_1),v_1\rangle \\\quad \quad \quad + (\del{x}_0 {-} l_1\hatmap{q_1}\xi_1)\cdot u_1\big) + \\
\vdots \\
\big({\mbar_n}\langle \del{v}_0 {-} \sum_{k=1}^n l_k(\hatmap{q}_k\dot{\xi_k}+\hatmap{\dot{q_k}}\xi_k),v_n\rangle \\ \quad\quad\quad+ (\del{x}_0 {-} \sum_{k=1}^n l_k\hatmap{q_k}\xi_k)\cdot u_n\big)
\end{pmatrix}dt = 0 \label{eq:append_derv1}
\end{align}}
Using the following simplifications in \eqref{eq:append_derv1},
\begin{gather*}
	\langle -l_k(\hatmap{q}_k\dot{\xi_k}+\hatmap{\dot{q_k}}\xi_k), v\rangle = l_k(\hatmap{q_k}v)\cdot\dot{\xi}_k+l_k(\hatmap{\dq_k}v)\cdot \xi_k,\\
	-l_k(\hatmap{q_k}\xi_k)\cdot u = l_k(\hatmap{q_k}u)\cdot \xi_k,
\end{gather*}
and regrouping the respective variations would result in,
{\small \begin{align}
	\bigint \begin{pmatrix}
	\del{v}_0\cdot \sum_{k=0}^n\mbar_kv_k+ \del{x}_0\cdot \sum_{k=0}^nu_k + \\
	\Big[\dot{\xi}_1\cdot \big(l_1\hatmap{q_1}\sum_{k=1}^{n}\mbar_kv_k\big) + \\ {\footnotesize \xi_1\cdot\big(l_1\hatmap{\dot{q}_1}\displaystyle\sum_{k=1}^{n}\mbar_kv_k{+}l_1\hatmap{q_1}\displaystyle\sum_{k{=}1}^{n}u_k \big) \Big]+}\\
	\vdots \\
	\Big[	\dot{\xi}_n\cdot(\mbar_nl_n(\hatmap{q_n}v_n)) +\\ \xi_n\cdot(\mbar_nl_n(\hatmap{\dq_n}v)+l_n(\hatmap{q_n}u_n))\Big]
	\end{pmatrix}dt{=}0.
\end{align}}
Integration by parts on the respective variation sets results in
{\small \begin{align}
	\bigint \begin{pmatrix}
	-\del{x}_0\cdot (\mbar_0\dot{v}_0+\mbar_1\dot{v}_1+\hdots \mbar_n\dot{v}_n)+\\ \del{x}_0\cdot \big(\sum_{k=0}^nu_k\big)+ \\
	-{\xi}_1{\cdot }\bigg(\cancel{l_1\hatmap{\dot{q}_1}\sum_{k=1}^{n}\mbar_kv_k}{+} l_1\hatmap{{q}_1}\displaystyle\sum_{k=1}^{n}\mbar_k\dot{v}_k\bigg)\\ {+}\xi_1{\cdot}\bigg(\cancel{l_1\hatmap{\dot{q}}_1\displaystyle\sum_{k=1}^{n}\mbar_kv_k}{+}l_1\hatmap{q}_1\sum_{k=1}^{n}u_k \bigg)\\
	\vdots\\
	-{\xi}_n\cdot \bigg( l_n\hatmap{{q}_n}\mbar_n\dot{v}_n\bigg) +\xi_n\cdot\bigg(l_n\hatmap{q_n}u_n \bigg)
	\end{pmatrix}dt{=}0,
\end{align}}
and finally,
{\small
\begin{align}
	\bigint \begin{pmatrix}
	\del{x}_0\cdot \big(\sum_{k=0}^n-\mbar_k\dot{v}_k + u_k\big)+ \\
	{\xi}_1{\cdot} l_1\hatmap{{q}_1}\bigg(\sum_{k=1}^{n}-\mbar_k\dot{v}_k+u_k \bigg)\\
	\vdots\\
	{\xi}_n\cdot \bigg(- l_n\hatmap{{q}_n}\mbar_n\dot{v}_n+l_n\hatmap{q_n}u_n \bigg)
	\end{pmatrix}dt{=}0.
\end{align}}
	
By principle of least action the above integral is valid $\forall \del{x}_0,\xi_i,t $ and thus, to ensure the above equation to be zero for all time we have, 
	\begin{align}
	\begin{matrix}
	\big(-(\mbar_0\dot{v}_0+\mbar_1\dot{v}_1+\hdots \mbar_n\dot{v}_n) +  \sum_{k=0}^nu_k\big) = 0 \\
	q_1{\times}, \bigg(-l_1\hatmap{{q}}_1\sum_{k=1}^{n}\mbar_k\dot{v}_k+l_1\hatmap{q}_1\sum_{k=1}^{n}u_k \bigg) = 0,\\
	\vdots  \\
	q_n\times \bigg(- l_n\hatmap{{q}_n}\mbar_n\dot{v}_n+l_n\hatmap{q_n}u_n \bigg) = 0.
	\end{matrix} 
	\end{align}
%
Expanding the $\dot{v}_i$ we have, 
{\small \begin{gather*}
	(\mbar_0\dot{v}_0+\mbar_1(\dot{v}_0+\sum_{k=1}^{1}l_k\ddq_k)+\hdots+ \mbar_n(\dot{v}_0 + \sum_{k=1}^{n}l_k\ddq_k) ) =  \sum_{k=0}^nu_k, \\
	l_1(\hatmap{{q}}_1)^2\big(\mbar_1(\dot{v}_0+\sum_{k=1}^{1}l_k\ddq_k)+\hdots\mbar_n(\dot{v}_0 + \sum_{k=1}^{n}l_k\ddq_k)\big){=}l_1(\hatmap{q}_1)^2\sum_{k=1}^{n}u_k,\\
	\vdots\\
	l_n(\hatmap{{q}}_n)^2\big( \mbar_n(\dot{v}_0 + \sum_{k=1}^{n}l_k\ddq_k)\big) = l_n(\hatmap{q}_n)^2\sum_{k=n}^{n}u_k.
\end{gather*}}
Simplifying the above equations using \eqref{eq:mass_var} and the following relations,
\begin{gather*}
\dot{q}  = \omega{ \times }q,\\
\ddot{q}= \dot{\omega}{\times} q + \omega{\times}\dot{q}  = \dot{\omega}{\times }q - \|\omega\|^2q \\
(\hatmap{q})^2\ddot{q} = q{\times }(q{\times }\ddot{q}) = (q\cdot \ddot{q})q -(q\cdot q)\ddot{q} = -(\dot{q}\cdot \dot{q})q -\ddot{q},
\end{gather*}
would result in \eqref{eq:dyn_q}.

\section{Differential flatness}
\label{sec:AppendixB}
In this section, we present the proof for differential-flatness stated in Lemma~\ref{lemma:diff}
\begin{pf} Illustration of the differential-flatness is shown in Figure~\ref{fig:flatness}. For the purpose of proving differential-flatness we redefine the dynamics of the system using tensions in the cable links as given below,
	\begin{align}
	\mbar_0\ddot{x}_0 & = T_1 -\mbar_0g\ez, \label{eq:flatdyn-1}\\
	\mbar_a\ddot{x}_a & = T_{a{+}1} - T_a -\mbar_ag\ez, \label{eq:flatdyn-2}  \\
	\mbar_b\ddot{x}_b & = T_{b+1} - T_b  -\mbar_ag\ez + f_bR_b\ez,\label{eq:flatdyn-3}\\
	\mbar_b\ddot{x}_n & = - T_n  -\mbar_ng\ez + f_nR_n\ez,\label{eq:flatdyn-4}
	\end{align}
	where $\forall a\in \inds\backslash\{0,\indi\}$ \emph{i.e.,} all the points excluding the starting point of the cable and those connected to the quadrotors and $~\forall b\in \indi\backslash\{n\}$ and the quadrotor attitude dynamics are as given in \eqref{eq:dyn_R}. Also note $n{\in}\indi$, \emph{i.e.,} end the cable is attached to the quadrotor. Number of inputs in the system are $4n_Q$ corresponding to the thrust and moment of the quadrotors. Number of flat outputs are $3$ (for position $x_0$) + $3(n_Q-1)$ ($3\text{ for each tension }T_{k{+}1}\forall k\in\indi\backslash\{n\})\text{ and }n_Q(\text{ for each quadrotor yaw }\psi_j)=4n_Q$. 
	
	Making use of the these dynamics we prove the flatness as follows.
	\begin{enumerate}[label=(\roman{*})]
		\item Given, $x_0$ is a flat-output and therefore we have the cable start position and its derivatives as shown,
		\begin{gather}
		 \{x_0, \dot{x}_0, \ddot{x}_0,x^{(3)}_0,\hdots ,x^{(2n+4)}_0\} \label{eq:flatproof-1}.
		\end{gather}
		\item Taking derivatives of \eqref{eq:flatdyn-1} and making use of \eqref{eq:flatproof-1} we have tension vector $T_1$ in the first link and its derivatives, 
		\begin{gather}
		\{T_1, \dot{T}_1,\ddot{T}_1, T^{(3)}_1\hdots, T^{(2n+2)}_1 \}. \label{eq:flatproof-2}
		\end{gather}
		\item Attitude of the first link is then determined from the tension vector in \eqref{eq:flatproof-2} as 
		\begin{gather}
		q_1 = T_1/\|T_1\| \label{eq:flatproof-q1}
		\end{gather}
		and its higher derivatives,
		\begin{gather}
		\{\dot{q}_1,\hdots, q^{(2n+2)}_1 \},\label{eq:flatproof-3}
		\end{gather} are computed by taking derivatives of \eqref{eq:flatproof-q1}  and using \eqref{eq:flatproof-2}.
		
		\item Position and its derivatives of the next link point-mass $\mbar_1$ is computed using \eqref{eq:link-positions}, 
		\begin{gather}
		\{ x_1,\dot x_1,\ddot x_1,\hdots, x_1^{(2n+2)}  \}. \label{eq:flatproof-4}
		\end{gather}
		\item Repeating the steps (ii)-(iv), we can compute the link attitudes, tensions and the positions iteratively till $x_b$. 
		\item Using \eqref{eq:flatdyn-3} and the fact that $T_{b+1}$ is a flat-output (note $ b\in \indi\backslash\{n\}$) we can compute the thrust in the quadrotor $f_bR_b\ez$.
		\item From $x_b$, $f_bR_b\ez$ and their derivatives, the quadrotor attitude, angular velocity $R_j,\Omega_j$ and moment $M_j$ can be computed as shown in [\cite{mellinger2011minimum}].
		\item Rest of the states and inputs for the \currsys~segments can be iteratively determined as described above. 
	\end{enumerate}
\end{pf}

Proof for Corollary~\ref{cor} is given below, 
\begin{pf}
 For tethered system we have $x_0\equiv0$ and $T_1$ is known since it-is a flat-output, i.e., steps (i)-(ii) (see \eqref{eq:flatproof-1}-\eqref{eq:flatproof-2}). Rest of the proof follows form Lemma~\ref{lemma:diff}. 
\end{pf}



\section{Variation-based Linearized Dynamics}
\label{sec:AppendixC}
Taking variations with respect to desired states for various states is as follows, 
\begin{gather}
\delta q_i = \hatmap{\xi}_iq_{id} = -\hatmap{q}_{id}\xi_i \\
\delta(\|\omega_i\|^2)  = \delta(\omega_i^\top \omega_i) = 2\omega_{id}^\top (\delta \omega_i)\\
\delta R_j = R_{jd}\hatmap{\eta}_j
\end{gather}
Taking variation on the first row of \eqref{eq:dyn_q},
{\small \begin{gather}
\delta\Bigg( M_{00}I_3\dot{v}_0 - \sum_{i=1}^{n}M_{0i}\hatmap{q}_{i}\dot\omega_i = \sum_{i=1}^{n}M_{0i}\|\omega_i\|^2q_i + \sum_{k=0}^{n}u_k \Bigg) 
\end{gather}
\begin{align}
M_{00}I_3\delta \dot v_0 - \sum_{i=1}^{n}M_{0i}\hatmap{q}_{id}\delta
	\dot \omega_i =\nonumber\\ \sum_{i=1}^{n}\Big(M_{0i}\big[(\hatmap{\dot{\omega}}_{id}-\|\omega_{id}\|^2I_3)\hatmap{q}_{id}\big]\xi_i + M_{0i}(2q_{id}\omega_{id}^\top)\delta\omega_i \Big) \nonumber \\ + \sum_{k=0}^{n}\big((\delta f_k)R_{kd}e_3\indicate{k} + f_{kd}\delta R_{kd}e_3\indicate{k}\big) \label{eq:variation1}
\end{align}}
taking variation on rest of the equations,
{\small \begin{align}
{\delta\Big(}M_{i0}\hatmap{q}_i\dot{v}_0+M_{ii}I_3\dot{\omega}_i - \sum_{j=1,j\neq i}^{n}M_{ij}\hatmap{q}_i\hatmap{q}_j\dot{\omega}_j =\nonumber \\  \sum_{k=1}^{n}(M_{ik}\|\omega_k\|^2\hatmap{q}_iq_k)+l_i(\hatmap{q}_i)\sum_{k=i}^{n}u_k{\Big)} 
\end{align}
\begin{align}
M_{i0}\big(\delta(\hatmap{q}_i)\dot{v}_{0d}+ \hatmap{q}_{id}\delta\dot{v}_0\big) + M_{ii}I_3\delta\dot{\omega}_i  \nonumber \\ - \sum_{j=1,j\neq i}^{n}M_{ij}\big[\delta(\hatmap{q}_{i})\hatmap{q}_{jd}\dot{\omega}_{jd} + \hatmap{q}_{id} \delta(\hatmap{q}_j)\dot{\omega}_{jd} + \hatmap{q}_{id}\hatmap{q}_{jd}\delta\dot{\omega}_j\big] \nonumber \\ 
= \sum_{k=1}^{n}(M_{ik}\big[2\hatmap{q}_{id}q_{kd}\omega^\top_{kd}\delta\omega_k + \|\omega_{kd}\|^2\delta(\hatmap{q}_i)q_{kd}+\nonumber \\ \|\omega_k\|^2\hatmap{q}_i\delta{q_{kd}}\big])+l_i\delta(\hatmap{q}_i)\sum_{k=i}^{n}u_k + l_i(\hatmap{q}_i)\sum_{k=i}^{n}\delta u_k
\end{align}

\begin{gather}
M_{i0}\hatmap{q}_{id}\delta\dot{v}_0 + M_{ii}I_3\delta\dot{\omega}_i  - \sum_{j=1,j\neq i}^{n}M_{ij}\big[ \hatmap{q}_{id}\hatmap{q}_{jd}\delta\dot{\omega}_j\big] \nonumber\\
= \Big[M_{io}\hatmap{\dot{v}_{0d}} -\sum_{j=1,j\neq i}^nM_{ij}\big(\hatmap{(\hatmap{q}_{jd})\dot{\omega}_{jd})} + \|\omega_{jd}\|^2\hatmap{q}_{jd} \big)-l_i{\big(\sum_{k=i}^n\hatmap{u}_k\big)} \Big](-\hatmap{q}_i){\xi}_i \nonumber \\
+ \sum_{j=1,j\neq i}^n\Big[M_{ij}\hatmap{q}_{id}\big(\hatmap{\dot\omega}_{jd} - \|\omega_{jd}\|^2I\big)\hatmap{q}_{jd}\Big]\xi_j + \sum_{j=1,j\neq i }^{n}M_{ij}\big[2\hatmap{q}_{id}q_{jd}\omega^\top_{jd}\big]\delta\omega_j \nonumber\\+ l_i(\hatmap{q}_i)\sum_{k=i}^{n}((\delta f_k)R_{kd}e_3\indicate{k} + f_{kd}\delta R_{kd}e_3\indicate{k}) \label{eq:variation2}
\end{gather}
\begin{gather}
M_{00}I_3\delta \dot v_0 - \sum_{i=1}^{n}M_{0i}\hatmap{q}_{id}\delta
	\dot \omega_i =\nonumber \\  \sum_{i=1}^{n}\Big(M_{0i}\big[(\hatmap{\dot{\omega}}_{id}-\|\omega_{id}\|^2I_3)\hatmap{q}_{id}\big]\xi_i + M_{0i}(2q_{id}\omega_{id}^\top)\delta\omega_i \Big) +\nonumber \\  \sum_{k{\in}\indi}\big((\delta f_k)R_{kd}e_3 + f_{kd}\delta R_{kd}e_3\big)
\end{gather}
\begin{gather}
M_{i0}\hatmap{q}_{id}\delta\dot{v}_0 + M_{ii}I_3\delta\dot{\omega}_i  - \sum_{j=1,j\neq i}^{n}M_{ij}\big[ \hatmap{q}_{id}\hatmap{q}_{jd}\delta\dot{\omega}_j\big] \nonumber\\
= \Big[M_{io}\hatmap{\dot{v}_{0d}} -\sum_{j=1,j\neq i}^nM_{ij}\big(\hatmap{(\hatmap{q}_{jd})\dot{\omega}_{jd})} +   \|\omega_{jd}\|^2\hatmap{q}_{jd} \big)-l_i{\big(\sum_{k=i}^n\hatmap{u}_k\big)} \Big](-\hatmap{q}_i){\xi}_i \nonumber \\
+ \sum_{j=1,j\neq i}^n\Big[M_{ij}\hatmap{q}_{id}\big(\hatmap{\dot\omega}_{jd} - \|\omega_{jd}\|^2I\big)\hatmap{q}_{jd}\Big]\xi_j +  \sum_{j=1,j\neq i }^{n}M_{ij}\big[2\hatmap{q}_{id}q_{jd}\omega^\top_{jd}\big]\delta\omega_j \nonumber\\+ l_i(\hatmap{q}_i)\sum_{k=i}^{n}((\delta f_k)R_{kd}e_3\indicate{k} + f_{kd}\delta R_{kd}e_3\indicate{k}) \label{eq:variation2}
\end{gather}}
From \eqref{eq:variation1} \& \eqref{eq:variation2}, we have,
\onecolumn

	\begin{gather}
	\begin{bmatrix}
	M_{00}I_3 & -\hatmap{q}_{1d}M_{01} & -\hatmap{q}_{2d}M_{02} & \hdots &  -\hatmap{q}_{nd}M_{0n} \\[2ex]
	M_{10}\hatmap{q}_1 & M_{11}I_3 & -M_{12}\hatmap{q}_{1d}\hatmap{q}_{2d} & \hdots & -M_{1n}\hatmap{q}_{1d}\hatmap{q}_{nd}\\[2ex]
	M_{20}\hatmap{q}_{2d} & -M_{21}\hatmap{q}_{2d}\hatmap{q}_{1d} & M_{22}I_3 & \hdots & -M_{2n}\hatmap{q}_{2d}\hatmap{q}_{nd}\\[2ex]
	\vdots  & \vdots & \vdots & \ddots & \vdots  \\[2ex]
	M_{n0}\hatmap{q}_{nd} & -M_{n1}\hatmap{q}_{nd}\hatmap{q}_{1d} & -M_{n2}\hatmap{q}_{nd}\hatmap{q}_{2d} & \hdots & M_{nn}I_3
	\end{bmatrix} \begin{bmatrix}
	\delta v_0 \\ \delta\dot\omega_1 \\ \vdots \\ \delta\dot\omega_n
	\end{bmatrix} \nonumber \\
	\begin{bmatrix}
	O & a_1 & \hdots & a_n & O & b_1 & \hdots & b_n & e_1 &  \hdots & e_{n_Q} & O &  \hdots & O \\
	O & c_{11} & \hdots & c_{1n} & O & d_{11} & \hdots & d_{1n} & f_{11}& & f_{1n_Q}& O & \hdots & O\\
	\vdots & \vdots & \ddots & \vdots & \vdots & \vdots & \ddots & \vdots & & \ddots & & \vdots & \ddots & \vdots \\
	O & c_{n1} & \hdots & c_{nn} & O & d_{n1} & \hdots & d_{nn}  & & & f_{n_Qn_Q} & O & \hdots & O
	\end{bmatrix} \begin{bmatrix}
	\delta x \\ \xi_1\\ \vdots \\ \xi_n \\ \delta v \\ \delta\omega_1 \\ \vdots \\ \delta\omega_n \\ \eta_{j1} \\ \vdots \\ \eta_{jn_Q} \\ \delta\Omega_{j1}  \\ \vdots \\ \delta\Omega_{jn_Q}
	\end{bmatrix} + G\del{\mathbf{ u}}, 
	\end{gather}
	where
	\begin{align}
	a_i &= M_{0i}\big[(\hatmap{\dot{\omega}}_{id}-\|\omega_{id}\|^2I_3)\hatmap{q}_{id}\\
	b_i &= M_{0i}(2q_{id}\omega_{id}^\top),\quad i=\{1,\hdots,n\}\\
	c_{ij} &= \begin{cases}
	\big[M_{io}\hatmap{\dot{v}_{0d}} -\sum_{j=1,j\neq i}^nM_{ij}\big(\hatmap{(\hatmap{q}_{jd})\dot{\omega}_{jd})} +\|\omega_{jd}\|^2\hatmap{q}_{jd} \big)-l_i{\big(\sum_{k=i}^n\hatmap{u}_k\big)} \big](-\hatmap{q}_i) &\quad i=j\\
	\big[M_{ij}\hatmap{q}_{id}\big(\hatmap{\dot\omega}_{jd} - \|\omega_{jd}\|^2I\big)\hatmap{q}_{jd}\big] & \quad i\neq j
	\end{cases}\\
	d_{ij} &= \begin{cases}
	O &\quad i=j\\
	M_{ij}\big[2\hatmap{q}_{id}q_{jd}\omega^\top_{jd}\big] &\quad i\neq j
	\end{cases}\\
	e_j & =  -f_{jd}R_{jd}\hatmap{e}_3,\quad j\in{\indi}\\
		f_{ij} & = \begin{cases} [] & if~j\notin \indi \\
	-(l_i\hatmap{q}_i)f_{jd}R_{jd}\hatmap{e_{3}}, &if ~j\in\indi, j\geq i \\
	O_{3,3} & if~ j\in \indi, j<i 
	\end{cases} \\
	g_j & = R_{jd}e_3\\
	h_{ij} & = \begin{cases} [] & if~j\notin \indi \\
	(l_i\hatmap{q}_i)R_{jd}{e_{3}}, &if ~j\in\indi, j\geq i \\
	O_{3,1} & if~ j\in \indi, j<i 
	\end{cases}
	\end{align}
	

\begin{tcolorbox}[title={Variation Linearized Dynamics}]
	\begin{align}
	\mathbf{\delta x}  &= [ \delta x ,~\xi_1,~\hdots,~ \xi_n,~\delta v,~\delta\omega_1,~\hdots, \delta\omega_n, \nonumber \\
	& \quad \quad\eta_{j1},~ \hdots,~ \eta_{jn_Q},~ \delta\Omega_{j1},~\hdots,~\delta\Omega_{jn_Q}]^\top \\
	\mathbf{\delta u} & = [f_{j1}, f_{j2}, \hdots, f_{nQ}, M_{j1}^\top , M_{j2}^\top,\hdots,M_{jn_Q}]^\top
	\end{align} 
	\begin{gather}
	\delta\dot{\mathbf{ x}} = \mathcal{A}\delta \mathbf{ x} + \mathcal{B}\delta\mathbf{u} \label{eq:var_lin_dyn_appendixC}
	\end{gather}
	where,{\small
		\begin{gather} 
		\mathcal{A} = \begin{bmatrix}
		0_{3,3} & 0_{3,3n}  & I_{3,3}  & 0_{3,3n}  & 0_{3,3n_Q}  & 0_{3,3n_Q}  \\
		0_{3n,3} & \alpha & 0_{3,3} & \beta & 0_{3,3n_Q}  & 0_{3,3n_Q} \\
		&&& M^{-1}F &  \\
		0_{3n_Q,3} & 0_{3n_Q,3n} & 0_{3n_Q,3} & 0_{3n_Q,3n} & \gamma & I_{3n_Q,3n_Q}\\
		0_{3n_Q,3} & 0_{3n_Q,3n} & 0_{3n_Q,3} & 0_{3n_Q,3n} & 0_{3n_Q,3n_Q} & \nu
		\end{bmatrix},\\[3ex]
		\alpha = \textit{blkdiag}[q_{1d}q_{1d}^\top\hatmap{\omega}_{1d},~ q_{2d}q_{2d}^\top\hatmap{\omega}_{2d}, ~\hdots,~ q_{nd}q_{nd}^\top\hatmap{\omega}_{nd}]\\
		\beta = \textit{blkdiag}[ \big(I_3{-}q_{1d}q_{1d}^\top\big), \big(I_3{-}q_{2d}q_{2d}^\top\big),\hdots, \big(I_3{-}q_{nd}q_{nd}^\top\big)]\\
		\gamma = \textit{blkdiag}[-\hatmap{\Omega}_{j1d},~-\hatmap{\Omega}_{j2d},\hdots,-\hatmap{\Omega}_{jn_Qd}]\\
		\nu = \textit{blkdiag}[J_1^{-1}(\hatmap{(J_1\Omega_{1d})}{-}\hatmap{\Omega}_{1d}J_1),..,J_1^{-1}(\hatmap{(J_n\Omega_{nd})}{-}\hatmap{\Omega}_{nd}J_n)]
		\end{gather}
		\begin{gather}
		\mathcal{B} = \begin{bmatrix}
		O_{3(n{+}1), 4n_Q} \\ M^{-1}G \\ 
		O_{3n_Q, 4n_Q} \\ 
		\begin{bmatrix}
		O_{3n_Q, n_Q} & \mu
		\end{bmatrix}
		\end{bmatrix},
		\mu = \textit{blkdiag}[J^{-1}_{j1}, \hdots, J^{-1}_{jn_Q}]
		\end{gather}}
	and {\small
		
		\begin{gather}
		M = \begin{bmatrix}
		M_{00}I_3 & -\hatmap{q}_{1d}M_{01} & \hdots &  -\hatmap{q}_{nd}M_{0n} \\[2ex]
		-M_{10}\hatmap{q}_{1d} & -M_{11}I_3  &{\scriptsize  \hdots} & M_{1n}\hatmap{q}_{1d}\hatmap{q}_{nd}\\[2ex]
		\vdots  & \vdots & {\scriptsize \ddots} & \vdots  \\[2ex]
		-M_{n0}\hatmap{q}_{nd} & M_{n1}\hatmap{q}_{nd}\hatmap{q}_{1d} &  {\scriptsize \hdots} & -M_{nn}I_3
		\end{bmatrix}\\
		F = \begin{bmatrix} 
		O_{3,3} & [a]_i & O_{3,3} & [b]_i & [e]_j & O_{3n_Q, 3n_Q} \\
		O_{3n,3}& [c]_{i,j} & O_{3n,3} & [d]_{i,j} & [f]_{ij}  & O_{3n_Q, 3n_Q}
		\end{bmatrix}\\
G = \begin{bmatrix}
\begin{bmatrix}
[g ]_i \\ [h]_{i,j}
\end{bmatrix} &  O_{(3(n+1), 3n_Q)}
\end{bmatrix}
		\end{gather}}
\end{tcolorbox}

%
%
%
%

\end{document}